\documentclass[conference]{IEEEtran}
\IEEEoverridecommandlockouts
\usepackage[space]{cite}
\usepackage{amsmath,amssymb,amsfonts}
\usepackage{float}
\usepackage{booktabs}
\usepackage{multirow}
\usepackage{algorithm}
\usepackage{algpseudocode}
\usepackage{graphicx}
\usepackage{textcomp}
\usepackage{hyperref}
\usepackage[table]{xcolor}
\usepackage[inline]{enumitem}

\newif\ifdraft
\drafttrue        

\usepackage{xcolor}

\begin{document}

\title{When to Ask for Help: Uncertainty-Triggered Aerial Refinement for Off-Road Navigation}
\author{Pranay Samineni, Adam Thorpe, John Chubick, Anand Balakrishnan, \\Christian Ellis, Swarat Chaudhuri, Ufuk Topcu}

\maketitle

\begin{abstract}
Heterogeneous air--ground teams can use aerial sensing to address the limited range and viewpoint of ground-vehicle sensors, but deciding when the UGV needs assistance and what information the UAV should collect remains a challenge.
This paper presents a planner-conditioned air--ground assistance method that evaluates how uncertain map information affects the UGV's current MPPI rollouts.
When navigation remains degraded, we score the same rollouts under resolved map hypotheses, select the map layer with the largest useful improvement, request the corresponding aerial measurement, fuse the returned information, and replan.
Unlike approaches driven by fixed support roles, task structure, or uncertainty alone, our method conditions assistance on whether resolving uncertainty improves predicted ground navigation.
Across 180 simulation trials on three off-road routes, the method achieves 98.9\% goal attainment while reducing mean UAV flight time by 87.7\% relative to continuous greedy aerial mapping.
Aggregated across the routes, each second of estimated UAV flight time saves 2.94 s of UGV navigation time.
Field experiments spanning approximately 2 km of UGV operation demonstrate uncertainty resolution and successful navigation recovery. Project page: \url{https://anonymous.4open.science/w/when-to-ask-for-help/}.
\end{abstract}


\section{Introduction}
Ground vehicles operating alone in outdoor, unstructured environments are limited by their sensing perspective, mobility, and onboard perception.
Ground-level sensors have limited range, suffer from occlusion, and can provide incomplete or ambiguous information about terrain, obstacles, and vehicle state.
An unmanned ground vehicle (UGV) may therefore fail to complete an otherwise achievable mission because it cannot obtain the information needed to navigate safely and reliably.
Heterogeneous robot teams address this by combining complementary sensing and mobility.
An unmanned aerial vehicle (UAV) can observe the environment and the ground vehicle from viewpoints unavailable to the UGV and provide information that improves the ground robot's navigation state.
However, UAV assistance is limited by flight endurance and cannot be used continuously throughout long missions.
The key problem is therefore deciding when the UGV needs assistance and what information the UAV should provide.

This paper presents a heterogeneous air--ground teaming approach for recovering degraded UGV navigation caused by incomplete or uncertain information in off-road environments.
The UGV maintains separate navigation map layers together with their associated uncertainty. 
When navigation degrades, we evaluate the current MPPI rollout population~\cite{williams2016aggressive} against counterfactual versions of these map layers and score how the resulting map changes affect predicted ground motion.
We then select the corresponding UAV sensing action for the map layer with the largest useful improvement.
The UAV collects the requested aerial information, such as an elevation map or semantic map, and communicates it to the UGV. The UGV fuses the information into its navigation maps, updates the planner representation, and replans.

\begin{figure}
    \centering
    \includegraphics[width=0.95\linewidth]{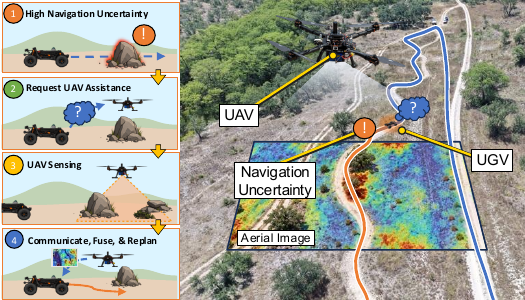}
    \caption{The presented planner-conditioned air-ground assistance method that enables a UGV to navigate highly uncertain off-road environments by requesting targeted aerial assistance to augment its own sensing.}
    \label{fig:hero-picture}
    \vspace{-.5em}
\end{figure}

Existing air-ground systems use heterogeneous sensing for cooperative localization, mapping, exploration, and navigation.
Many approaches assign the UAV a fixed support role, invoke aerial sensing according to predefined task logic, or trigger additional sensing from uncertainty in a perception output \cite{garzon2013aerialground}.
These sensing criteria do not directly determine whether the uncertainty affects the UGV's current navigation decision.
Uncertainty can be large in regions that have no effect on the current route, while localized uncertainty can prevent the UGV from identifying or executing a safe path.
We therefore evaluate uncertainty through its effect on ground navigation and request UAV assistance only when resolving uncertainty is predicted to improve the UGV's ability to make progress.

Our main contribution, illustrated in Figure~\ref{fig:hero-picture}, is a planner-conditioned air--ground assistance method that determines when the UAV should sense by evaluating resolved map hypotheses against current UGV rollouts and selecting an aerial sensing action based on the change in navigation score.
Across 180 simulation trials, our method achieves 98.9\% goal attainment with 87.7\% less mean UAV flight time than continuous greedy mapping.
Across three routes, each estimated UAV flight second saves $2.94\,\mathrm{s}$ of UGV navigation time.
Field experiments spanning approximately 2~km of autonomous UGV operation through unstructured routes demonstrate repeated resolution of map uncertainty and resumed navigation following aerial assistance.

\begin{figure*}[t]
    \centering
    \includegraphics[width=0.95\linewidth]{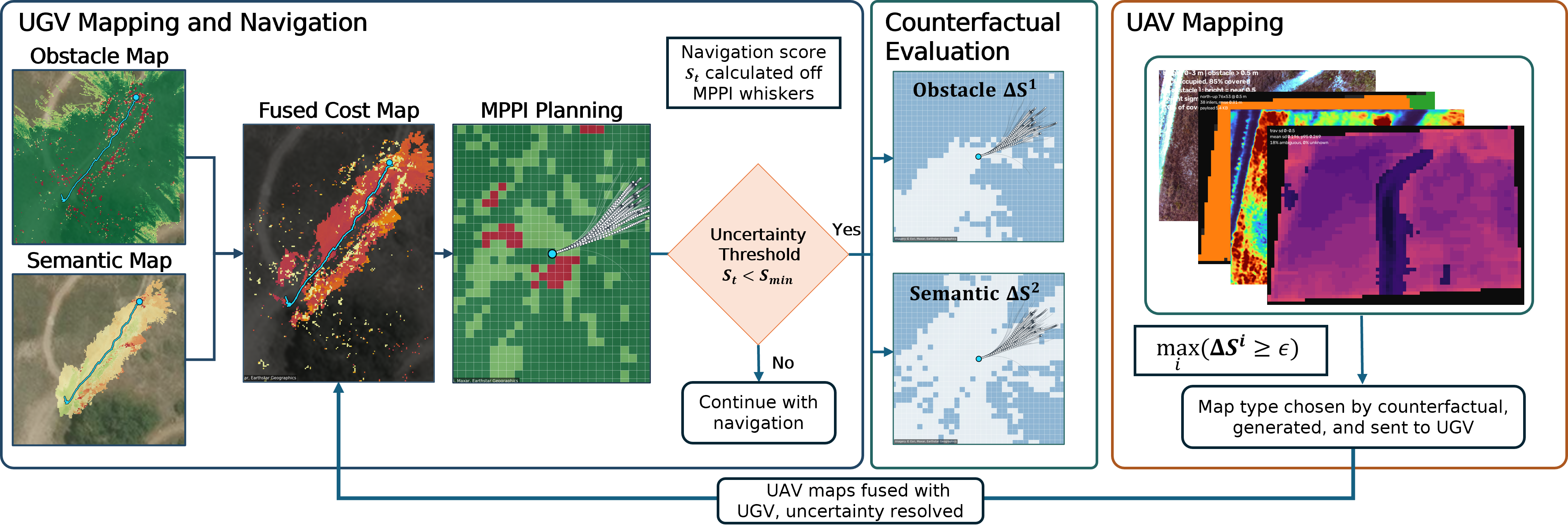}
    \caption{The UGV consumes LiDAR, RGB-D, and IMU data to build layers of maps (obstacles, semantic traversability) that serve as a basis for a fused costmap for motion planning. During planning time, while the projected MPPI plans stay below an uncertainty threshold, the nominal MPPI plan is used.
    Otherwise, \emph{counterfactual} diagnosis is used to request the best aerial map that would resolve the uncertainty.}
    \label{fig:autonomy-pipeline}
\end{figure*}
\section{Related Work}

\textit{Field Autonomy.}
Autonomous navigation in unstructured outdoor environments requires a ground robot to reason about obstacles, terrain, vehicle state, and long-range route structure from incomplete onboard observations. Early off-road autonomy systems identified limited sensing lookahead and terrain interpretation as fundamental constraints on ground navigation \cite{howard2007traversability,bajracharya2009lagr}. Recent work has improved long-range planning through learned cost maps \cite{trailblazer2025}, terrain--vehicle interaction models \cite{anynav2025}, and increasingly capable local planners evaluated under physical field conditions \cite{majhor2025muons}. These approaches improve the information available to the ground-navigation stack, but the sensing geometry and mobility of the UGV continue to limit what can be observed from the ground. We consider failures that arise when these information limitations prevent an otherwise capable navigation stack from identifying or executing useful motion.

\textit{Air--Ground Teaming.}
Air--ground teams use the complementary sensing and mobility of aerial and ground platforms to extend the capabilities of either platform operating independently.
Early systems used UAVs as forward scouts to detect hazards beyond the sensing horizon of an off-road UGV \cite{stentz2002flyingeye,delmerico2017activeexploration,zhang2022fastactive}, while subsequent work has considered heterogeneous teams for perception, exploration, localization, mapping, and navigation \cite{zhou2026survey}. 
Recent approaches increasingly adapt aerial sensing to the state of the task.
ColAG triggers aerial assistance when perception-limited UGVs predict collision risks arising from localization uncertainty or unknown regions along their planned trajectories \cite{li2024colag}.
PACE proactively schedules aerial perception at information-rich frontiers under timing constraints imposed by the motion of the UGV \cite{jiao2026pace}. While some prior approaches gather aerial information before deploying a UGV \cite{delmerico2017activeexploration,rockenbauer2024mars}, our approach selectively invokes aerial assistance during ongoing ground navigation based on its predicted benefit to ground motion, enabling more efficient UAV utilization.

\textit{Collaborative Navigation and Mapping.}
A substantial body of air--ground research uses aerial sensing to improve the information available to a ground-navigation system.
UAV imagery has been used to construct obstacle maps and guide UGV motion in large outdoor environments \cite{kim2013collaborative,garzon2013aerialground}, while cooperative mapping systems fuse aerial and ground observations to improve environmental coverage \cite{kim2019cooperativemapping}.
Related approaches use inter-platform measurements to improve localization, including active motion strategies for positioning geometry \cite{sivaneri2017uavugv,sivaneri2018flighttest}.

Recent work couples aerial perception more directly to UGV navigation.
Fortin et al.\ estimate off-road terrain properties from aerial imagery to construct traversability information for subsequent UGV path planning \cite{fortin2025uavterrain}.
Counterfactual reasoning has also been applied to navigation: CREStE rewards navigation aligned with human preferences using counterfactual trajectory annotations \cite{zhang2025creste}, and CoINS reasons about hypothetical object removal to decide when physical intervention is needed to create a traversable path \cite{zhou2026coins}. 
Our approach instead evaluates hypothetical resolutions of map uncertainty by their effect on the ground planner, an optimistic form of value of information~\cite{howard1966information} applied to deciding when a robot should ask for help~\cite{tellex2014asking}, with a UAV rather than a human as the helper.
The UGV scores its current planner rollouts under counterfactual resolutions of each uncertain map layer and requests aerial sensing only when a resolution yields sufficient improvement in predicted ground motion.
This lets one navigation loop select among multiple aerial sensing roles and leaves the UAV idle when unresolved uncertainty does not constrain the current decision.


\section{Planner-Conditioned Aerial Assistance}
Aerial sensing can provide information that is unavailable from the UGV's ground-level viewpoint, but uncertainty alone does not determine when that sensing is useful.
Large uncertainty may occur in regions that do not affect the current navigation problem, while a small uncertain region may prevent the planner from finding viable motion.
The assistance decision should therefore depend on whether unresolved information affects the motions currently available to the UGV and whether resolving that could change the navigation outcome.

We condition assistance on the current planning state.
As seen in Figure~\ref{fig:autonomy-pipeline},
the UGV monitors the current MPPI rollouts~\cite{williams2016aggressive} and detects when navigation degrades.
When degradation is detected, the system evaluates the uncertain map sources against the same rollout population.
For each uncertainty source, it constructs a plausible resolved map, rescores the current rollouts under that counterfactual, and measures the resulting change in navigation score.
If a source exceeds the intervention threshold, meaning it would improve the UGV route if it were resolved, the UGV sends the corresponding sensing request to the UAV.
The UAV acquires the requested information, returns the measurement, and the UGV updates its navigation representation and replans.
If no source produces sufficient improvement, no aerial request is made.

\textit{Problem Formulation.}
We consider a UGV navigating an outdoor, unstructured environment with a receding-horizon local planner.
At time $t$, let $x_t$ denote the UGV state used by the planner and $g_t$ the current local planning goal.
The UGV maintains $L$ navigation map layers $M_t=\{m_t^1,\ldots,m_t^L\}$, where each layer $m_t^i$ contains information used for navigation, such as obstacle occupancy, terrain semantics, or geometric traversability.
Each map layer $m_t^i$ has an associated uncertainty representation $u_t^i$, collected as $U_t=\{u_t^1,\ldots,u_t^L\}$.
Different map layers may represent uncertainty differently, and we do not assume a common uncertainty model across all map layers.

The navigation system combines the map layers into a planner cost representation $C_t=F(M_t)$, where $F$ denotes the map-fusion and cost-generation procedure.
The uncertainty representations $U_t$ are retained separately for assistance diagnosis.
Given the state $x_t$, goal $g_t$, and planner representation $C_t$, the local planner $\Pi$ generates $K$ candidate trajectories,
\begin{equation}
T_t=\Pi(x_t,g_t,C_t)=\{\tau_t^{(1)},\ldots,\tau_t^{(K)}\}.
\label{eq:rollout_population}
\end{equation}
The rollout population $T_t$ contains the motions evaluated by the planner at time $t$.
We summarize the current planner state with a scalar navigation score $S_t=S(T_t,C_t)$, which indicates whether the current rollout population supports useful ground motion.
A degraded value of $S_t$ can initiate diagnosis, but does not by itself cause an aerial request.

Let $\mathcal{A}=\{a_1,\ldots,a_N\}$ denote the aerial sensing actions. At an assistance decision, the system selects $a_t^\star\in\mathcal{A}\cup\{\emptyset\}$, where $a_t^\star=\emptyset$ means that no aerial assistance is requested.

The problem is to determine whether prolonged uncertainty in any map layer affecting the current rollout population $T_t$ can be resolved in a way that improves the navigation score $S_t$. If so, the goal is to select the aerial sensing action $a_t^\star$ associated with the map layer whose resolution produces the largest predicted improvement.

\textit{Navigation Degradation Detection.}
We begin assistance diagnosis only when degraded navigation persists. Let $\Gamma_t\in\{0,1\}$ denote the degradation trigger,
\begin{equation}
\Gamma_t=
\begin{cases}
1, & S_{t-j}<S_{\min}\text{ for }j=0,\ldots,N_p-1,\\
0, & \text{otherwise},
\end{cases}
\label{eq:degradation_trigger}
\end{equation}
where $S_{\min}$ is the minimum acceptable navigation score and $N_p$ is the number of consecutive planning cycles required to trigger diagnosis. The persistence requirement prevents temporary changes from starting the assistance procedure.

When $\Gamma_t=1$, we pass the current map layers $M_t$, uncertainty representations $U_t$, and rollout population $T_t$ to the counterfactual diagnosis. At this stage, the system has only established that navigation remains degraded. The following counterfactual evaluation determines whether resolving any map uncertainty is expected to improve the navigation score.

\textit{Planner-Conditioned Counterfactual Diagnosis.}
A degraded navigation score does not by itself show that additional sensing will help. We therefore evaluate uncertainty in each map layer against the current rollout population $T_t$.
For map layer $m_t^i$ with uncertainty representation $u_t^i$, let $\mathcal{R}_i$ denote a set of plausible resolved hypotheses. Each $r\in\mathcal{R}_i$ defines a counterfactual map layer $\tilde{m}_t^{i,r}$ by resolving the uncertain portion of $m_t^i$ that affects the current rollouts. All other map layers, the current state $x_t$, the local goal $g_t$, and the rollout population $T_t$ remain unchanged. For an occupancy layer, for example, $\mathcal{R}_i$ may contain free and occupied hypotheses over uncertain cells intersected by the rollout footprints.

Replacing $m_t^i$ with $\tilde{m}_t^{i,r}$ gives the counterfactual map set $\tilde{M}_t^{i,r}=\{m_t^1,\ldots,\tilde{m}_t^{i,r},\ldots,m_t^L\}$ and planner representation $\tilde{C}_t^{i,r}=F(\tilde{M}_t^{i,r})$. We then score the same rollout population as $\tilde{S}_t^{i,r}=S(T_t,\tilde{C}_t^{i,r})$. Holding $T_t$ fixed isolates the effect of the modified map information from changes caused by replanning.

For map layer $i$, we define the counterfactual navigation improvement, $\Delta S_t^i$, as the largest improvement in navigation score across the resolved hypotheses evaluated for map layer $m_t^i$ and is the quantity used for assistance selection
\begin{equation}
\Delta S_t^i=\max_{r\in\mathcal{R}_i}\tilde{S}_t^{i,r}-S_t.
\label{eq:counterfactual_improvement}
\end{equation}


\textit{Assistance Selection.}
Let $\epsilon$ denote the minimum counterfactual improvement required to justify an aerial request. We first consider only map layers satisfying $\Delta S_t^i\ge\epsilon$, then select the layer with the largest counterfactual improvement,
\begin{equation}
i^\star=\arg\max_i \Delta S_t^i.
\label{eq:selected_source}
\end{equation}
If no map layer satisfies $\Delta S_t^i\ge\epsilon$, no aerial request is issued.

Each map layer is associated with a sensing action that can reduce its uncertainty. We denote this mapping by $g$ and request $a_t^\star=g(i^\star)$. When the sensing action requires a spatial target, the request also includes the uncertain map region used in the counterfactual evaluation.

\textit{Aerial Refinement and Replanning.}
For the selected action $a_t^\star$, let $y_t$ denote the measurement returned by the UAV. We update the affected map and its uncertainty 
using an action-dependent refinement operator,
\begin{equation}
(M_t',U_t')=\Phi_{a_t^\star}(M_t,U_t,y_t).
\label{eq:aerial_refinement}
\end{equation}
Depending on the sensing action, $\Phi_{a_t^\star}$ may replace uncertain map values, fuse the aerial measurement with existing ground observations, or update the uncertainty associated with a map.

The UGV then rebuilds the planner representation as $C_t'=F(M_t')$. Using the current state $x_t$ and local goal $g_t$, the planner generates a new rollout population $T_t'=\Pi(x_t,g_t,C_t')$ and computes the updated navigation score $S_t'=S(T_t',C_t')$.

The UGV then reevaluates the degradation condition using the updated planning state. If the degradation condition clears, the UGV resumes nominal receding-horizon control. If degradation persists, we repeat the diagnosis using the updated maps $M_t'$, uncertainty representations $U_t'$, and rollout population $T_t'$. This allows later assistance requests to address uncertainty that remains after the first aerial measurement.

\textit{Alternative Diagnostics and Non-Map Uncertainty.}
Not all map representations admit simple map counterfactuals. Discrete maps such as occupancy grids can often be modified by assigning uncertain cells to alternative states. This is less natural for continuous maps such as elevation, roughness, or geometric cost maps. For these sources, we instead measure how much uncertain information intersects the current planning problem. For source $i$, let $X_t^i$ denote the uncertain region of map layer $m_t^i$ intersected by the current rollout population $T_t$, and let $\mathcal{F}(T_t)$ denote the set of map cells intersected by the swept rollout footprints. We define the fraction of the rollout footprint intersecting uncertain information from source $i$ as $R_t^i=|X_t^i|/|\mathcal{F}(T_t)|$, which measures the exposure of the current candidate motions to uncertainty from that source.

The same assistance logic could also be applied to non-map uncertainty. For example, localization drift can become important during long GPS-denied operation even when the local environment maps remain usable. Localization uncertainty could therefore be treated as another information source with its own diagnostic and sensing action. If that diagnostic indicates that localization is limiting navigation, the UAV could provide an external position measurement $y_t^{\mathrm{loc}}$ as an additional observation to the UGV's state estimator.

\section{System Implementation}
\label{sec:system_implementation}



\textit{UGV Sensing and Planning.}
LiDAR onboard the UGV is used to build a probabilistic obstacle occupancy grid, while a front-facing RGB-D camera provides terrain and traversability information from semantic segmentation. The obstacle and semantic maps use $0.25$~m cells over a $40\times40$~m local region. Obstacle occupancy stores per-cell occupancy probability and evidence statistics, terrain quantities store means and variances, and semantic cost is computed from the per-cell class distribution with corresponding uncertainty.


Given a route consisting of a sequence of GPS anchors,
the MPPI controller samples $K=128$ candidate control sequences over a $4.0$~s horizon and evaluates them using continuous route tracking, speed, control, obstacle, and active terrain or semantic costs. The resulting rollout trajectories are retained for assistance diagnosis.
For rollout $k$, let $q_{k,j}$ denote the maximum occupancy probability intersected by the vehicle footprint over trajectory segment $j$. We define the experimental navigation score as
$S_t=
\frac{1}{K}
\sum_{k=1}^{K}
\boldsymbol{1}_{k}^{\mathrm{useful}}
\prod_j(1-q_{k,j})$,
where $\boldsymbol{1}_{k}^{\mathrm{useful}}=1$ when the rollout remains inside the evaluated map and produces at least $0.25$~m of terminal displacement. A confirmed hard collision sets the rollout survival to zero. We declare navigation degraded when $S_t<0.20$. For obstacle assistance, this condition must persist for $2.0$~s before the counterfactual diagnosis is evaluated.

\textit{Assistance Diagnostic Implementation.}
We implement obstacle uncertainty using the counterfactual test in \eqref{eq:counterfactual_improvement}. At each diagnosis event, we hold the current MPPI rollout population $T_t$ fixed and identify uncertain occupancy cells intersected by the swept rollout footprints. We evaluate three occupancy interpretations: the current map, a free-space hypothesis in which the intersected uncertain cells are assigned occupancy probability $0.05$, and an occupied-space hypothesis in which they are assigned occupancy probability $0.95$. Known occupancy cells are left unchanged.

We score the same rollout population under each interpretation to obtain $S_t^{\mathrm{base}}$, $S_t^{\mathrm{free}}$, and $S_t^{\mathrm{occupied}}$. We then compute
\begin{equation}
\Delta S_t^{\mathrm{obs}}
=
\max(S_t^{\mathrm{free}},S_t^{\mathrm{occupied}})
-
S_t^{\mathrm{base}},
\label{eq:obstacle_counterfactual}
\end{equation}
and the separation between the two resolved hypotheses,
$D_t^{\mathrm{obs}}
=
|S_t^{\mathrm{free}}-S_t^{\mathrm{occupied}}|$.
The rollout scores use the obstacle survival or viability metric rather than rerunning the full MPPI optimization. An obstacle request is issued only when the counterfactual test also satisfies the implemented obstacle-crossing, uncertainty-exposure, viability, separation, and persistence conditions.

We implement semantic and geometric uncertainty using rollout-conditioned exposure rather than explicit resolved-map hypotheses. For each map layer $m_t^i$, we project the current rollout footprints into its uncertainty grid and compute the exposure $R_t^i=|X_t^i|/|\mathcal{F}(T_t)|$. We also record the uncertainty variance, evidence strength, and fraction of rollouts that intersect uncertain cells. A semantic request can be issued when this exposure persists above its source-specific threshold and the semantic layer contributes to the nominal MPPI cost.


\begin{table*}[!t]
\centering
\caption{Route-level and aggregate comparison over 30 runs in each direction.
Time, path, and intervention values are means over trials.
Route-level entries are written as mean\,$\pm$\,one standard deviation; the aggregate rows report means only.
Total mission time includes sequential UAV mapping time.}
\label{tab:route_policy_comparison}
\resizebox{\textwidth}{!}{%
\begin{tabular}{llrrrrrr}
\toprule
Route & Policy & Goal reached & Interventions &
UGV time (s) & UAV flight (s) &
Total mission time (s) & Path (m) \\

\midrule
Route-1 (EASY)
& Ground Only
& \textbf{60/60} & 0.62\,{\footnotesize$\pm$\,0.49} & 127.8\,{\footnotesize$\pm$\,33.2} & --- & 127.8\,{\footnotesize$\pm$\,33.2} & 131.7\,{\footnotesize$\pm$\,32.9} \\
\rowcolor{gray!12}
& UAV Counterfactual$^\dagger$
& \textbf{60/60} & \textbf{0.30}\,{\footnotesize$\pm$\,0.46} & 97.0\,{\footnotesize$\pm$\,18.7} & \textbf{11.3}\,{\footnotesize$\pm$\,4.9} & 108.4\,{\footnotesize$\pm$\,19.5} & 113.0\,{\footnotesize$\pm$\,11.4} \\
& Greedy
& 59/60 & 0.47\,{\footnotesize$\pm$\,1.33} & \textbf{77.7}\,{\footnotesize$\pm$\,14.0} & 77.7\,{\footnotesize$\pm$\,14.0} & \textbf{77.7}\,{\footnotesize$\pm$\,14.0} & \textbf{97.8}\,{\footnotesize$\pm$\,2.1} \\
& Explore then Drive
& \textbf{60/60} & 0.32\,{\footnotesize$\pm$\,0.47} & 79.3\,{\footnotesize$\pm$\,14.9} & 49.6\,{\footnotesize$\pm$\,0.0} & 129.0\,{\footnotesize$\pm$\,14.9} & 98.3\,{\footnotesize$\pm$\,2.0} \\

\midrule
Route-2 (MEDIUM)
& Ground Only
& 59/60 & \textbf{1.28}\,{\footnotesize$\pm$\,0.85} & 234.8\,{\footnotesize$\pm$\,79.6} & --- & 234.8\,{\footnotesize$\pm$\,79.6} & 267.1\,{\footnotesize$\pm$\,76.8} \\
\rowcolor{gray!12}
& UAV Counterfactual$^\dagger$
& 58/60 & 1.42\,{\footnotesize$\pm$\,2.35} & 205.7\,{\footnotesize$\pm$\,47.7} & \textbf{17.1}\,{\footnotesize$\pm$\,7.4} & 222.7\,{\footnotesize$\pm$\,51.3} & 235.7\,{\footnotesize$\pm$\,43.4} \\
& Greedy
& \textbf{60/60} & 1.47\,{\footnotesize$\pm$\,0.60} & \textbf{189.4}\,{\footnotesize$\pm$\,21.5} & 189.4\,{\footnotesize$\pm$\,21.5} & \textbf{189.4}\,{\footnotesize$\pm$\,21.5} & 216.2\,{\footnotesize$\pm$\,3.7} \\
& Explore then Drive
& \textbf{60/60} & 1.45\,{\footnotesize$\pm$\,0.72} & 190.6\,{\footnotesize$\pm$\,23.3} & 86.6\,{\footnotesize$\pm$\,0.0} & 277.2\,{\footnotesize$\pm$\,23.3} & \textbf{216.1}\,{\footnotesize$\pm$\,3.3} \\
\midrule
Route-3 (HARD)
& Ground Only
& 48/60 & 5.32\,{\footnotesize$\pm$\,1.72} & 782.1\,{\footnotesize$\pm$\,185.4} & --- & 782.1\,{\footnotesize$\pm$\,185.4} & 848.7\,{\footnotesize$\pm$\,260.4} \\
\rowcolor{gray!12}
& UAV Counterfactual$^\dagger$
& \textbf{60/60} & \textbf{2.87}\,{\footnotesize$\pm$\,1.29} & 567.8\,{\footnotesize$\pm$\,62.0} & \textbf{65.7}\,{\footnotesize$\pm$\,12.0} & 633.4\,{\footnotesize$\pm$\,66.6} & 629.8\,{\footnotesize$\pm$\,36.5} \\
& Greedy
& \textbf{60/60} & 2.90\,{\footnotesize$\pm$\,1.07} & 498.2\,{\footnotesize$\pm$\,49.1} & 498.2\,{\footnotesize$\pm$\,49.1} & \textbf{498.2}\,{\footnotesize$\pm$\,49.1} & 570.6\,{\footnotesize$\pm$\,5.6} \\
& Explore then Drive
& \textbf{60/60} & 2.93\,{\footnotesize$\pm$\,1.21} & \textbf{496.0}\,{\footnotesize$\pm$\,45.2} & 248.6\,{\footnotesize$\pm$\,0.0} & 744.6\,{\footnotesize$\pm$\,45.2} & \textbf{570.0}\,{\footnotesize$\pm$\,7.4} \\

\midrule
\multicolumn{8}{l}{\textbf{Aggregate Statistics Across Runs}} \\
\cmidrule(lr){1-8}

& Ground Only
& 167/180 (92.8\%) & 2.41 & 353.7 & --- & 353.7 & 385.6 \\
\rowcolor{gray!12}
& UAV Counterfactual$^\dagger$
& 178/180 (98.9\%) & \textbf{1.53} & 291.1 & \textbf{31.5} & 322.6 & 327.2 \\
& Greedy
& 179/180 (99.4\%) & 1.61 & 256.1 & 256.1 & \textbf{256.1} & 295.9 \\
& Explore then Drive
& \textbf{180/180 (100\%)} & 1.57 & \textbf{255.3} & 128.3 & 383.6 & \textbf{294.8} \\
\bottomrule
\end{tabular}%
}
\end{table*}

\textit{UAV Assistance and Measurement Fusion.}
UAV assistance is coordinated through an asynchronous map-request service. Each request specifies a region of interest and the required aerial action: a canopy-derived obstacle occupancy map or a semantic segmentation-derived traversability estimate.
For each request, the UAV captures an image of the desired ROI, 
and packages a response containing the geographically registered outputs of the requested mapping pipeline.

Returned maps are registered against the local inertial frame and fused into the corresponding UGV layers. Each layer maintains an estimate, variance, effective observation support, and age. Occupancy uses Bernoulli evidence fusion, with source contributions limited by their support and variance. Semantic costs use inverse-variance-weighted Gaussian fusion. 
After fusing the map layers, the UGV waits for a fixed fusion-settling interval before releasing the navigation hold. MPPI then evaluates new rollouts using the updated layers and continues with the navigation task.

\section{Simulation Evaluation and Results}
\label{sec:simulation_results}

We evaluate the assistance strategy in a Gazebo simulation environment using three unique navigation routes. The simulation evaluation addresses mission performance, aerial resource use, and whether the selected sensing action resolves information that constrains UGV navigation.
\begin{figure}[!b]
    \centering
    \includegraphics[width=0.9\linewidth]{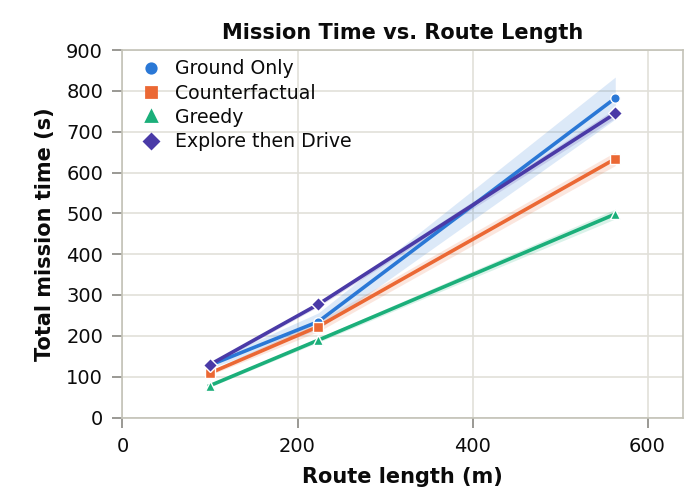}
    \caption{{Total mission time vs.\ route length. Relative to Ground Only (paired): Counterfactual 10.4\% faster, Greedy 38.3\% faster, Explore-then-Drive 10.9\% slower. Greedy's flight is concurrent and adds no mission time; Explore-then-Drive surveys before the rover moves. Bands: bootstrap 95\% CI.}
    \label{fig:mission_vs_route}}
\end{figure}

\textit{Simulation Setup and Evaluation Protocol.}
The simulation uses Gazebo Harmonic with a four-wheel drive, front-steered Ackermann rover, a 16-channel LiDAR, and front-facing depth camera for perception. A $512\,\mathrm{m} \times 512\,\mathrm{m}$ scene was constructed with LiDAR-derived elevation GeoTIFFs from the USGS 3D Elevation dataset spanning the same area as our physical unstructured experiment site \cite{sugarbaker2014usgs3d}. Grass, bushes, and trees are procedurally generated from hand-painted spatial masks, with seeded variation in position, scale, and orientation.

\begin{figure}[!b]
    \centering
    \includegraphics[width=0.9\linewidth]{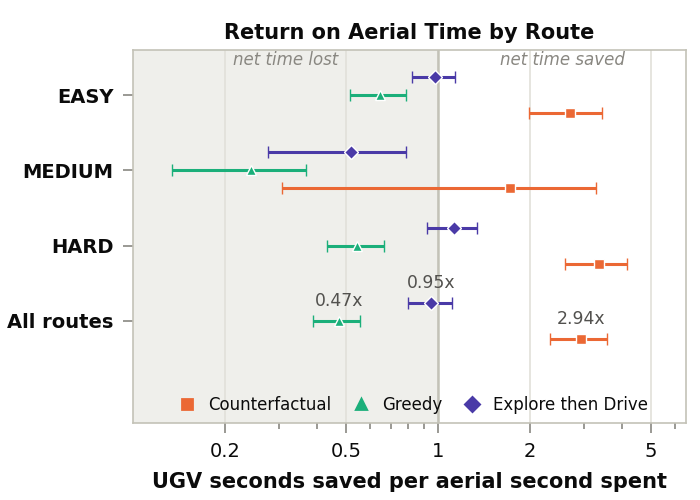}
    \caption{UGV navigation time saved for each aerial second spent, referenced off the matching Ground Only trials. Aggregate ratios are 2.94 for Counterfactual, 0.95 for Explore then Drive, and 0.47 for Greedy. The horizontal line marks one second of UGV time saved per aerial second. Bars: 95\% bootstrap CI.\label{fig:return_on_flight}}
\end{figure}

The simulation adapts the autonomy implementation from Section~\ref{sec:system_implementation}, using ground truth for localization and aerial maps.
UAV flight time is calculated analytically from the aerial maps required by each assistance strategy, allowing for a representative comparison of aerial resources without simulating actual UAV flight dynamics.
Along with metrics for goal completion, total mission time, and UAV flight time, we also measure the number of interventions performed to recover the UGV from a stuck state. 


Three routes of $99\,\mathrm{m}$, $224\,\mathrm{m}$, and $563\,\mathrm{m}$, modeled after real paths at the field site, are traversed in both directions and include elevation changes, vegetation, and physical obstacles.
These routes represent EASY, MEDIUM, HARD outdoor navigation tasks.
We evaluate four navigation policies:
\begin{enumerate*}[label=(\arabic*)]
\item
\textbf{Ground Only}, which uses onboard sensing without UAV assistance; 
\item
\textbf{UAV Counterfactual}, our proposed method, which requests targeted maps when counterfactual analysis predicts improved navigation, making UAV requests only when necessary; 
\item
\textbf{Greedy}, which continuously maps during UGV navigation; and 
\item
\textbf{Explore then Drive}, which maps the full route before UGV navigation begins.
\end{enumerate*}

\vspace{\baselineskip}
\textit{Mission Performance and Aerial Resource Use.}
Table~\ref{tab:route_policy_comparison} compares policies over 30 seeds per directional scenario. Trials are matched by route, direction, and random seed, with vehicle, perception, and planner parameters fixed across strategies.

UAV Counterfactual reaches the goal in 178/180 trials (98.9\%), compared with 167 (92.8\%) for Ground Only, 179 (99.4\%) for Greedy, and 180 (100\%) for Explore then Drive, supporting the contribution of aerial mapping to reliable navigation.
Relative to Ground Only, counterfactual assistance reduces mean UGV navigation time by 17.7\% ($353.7\,\mathrm{s}$ to $291.1\,\mathrm{s}$) and traveled distance by 15.1\% ($385.6\,\mathrm{m}$ to $327.2\,\mathrm{m}$). Figure~\ref{fig:mission_vs_route} shows total mission time against route length: in paired comparisons with Ground Only, counterfactual assistance completes missions 10.4\% faster, while Explore then Drive is 10.9\% slower due to its initial full-route survey.

On the HARD route, mean interventions decrease from 5.32 to 2.87 per trial and goal attainment increases from 48/60 to 60/60. The observed standard deviations of UGV time and distance also decrease from $185.4$ to $62.0\,\mathrm{s}$ and from $260.4$ to $36.5\,\mathrm{m}$, respectively, suggesting more consistent navigation in the reported trials.

As seen in Figure~\ref{fig:return_on_flight}, aggregated across the three routes, each second of estimated UAV flight time is associated with $2.94\,\mathrm{s}$ of UGV navigation time saved relative to Ground Only. This result quantifies the navigation benefit obtained from selectively acquiring aerial information using the proposed counterfactual method.

The proposed method maintains high goal attainment while substantially reducing aerial resource use. Among trials that reach the goal, mean estimated UAV flight time is $31.5\,\mathrm{s}$, an 87.7\% reduction relative to Greedy and a 75.4\% reduction relative to Explore then Drive. These savings accompany just one and two fewer goal-reaching trials out of 180, respectively. Mean total mission time is also 15.9\% lower than Explore then Drive ($322.6\,\mathrm{s}$ versus $383.6\,\mathrm{s}$). Although Greedy achieves the shortest mean mission time through concurrent mapping, it requires approximately 8.1 times as much UAV flight time. Together, these results show that counterfactual assistance retains the reliability benefits of aerial mapping with substantially lower aerial resource demand. 
Reported time, distance, and flight-time means are conditioned on goal attainment and therefore reflect different subsets of trials across policies.

\begin{figure}[!t]
    \centering
    \includegraphics[width=0.85\linewidth]{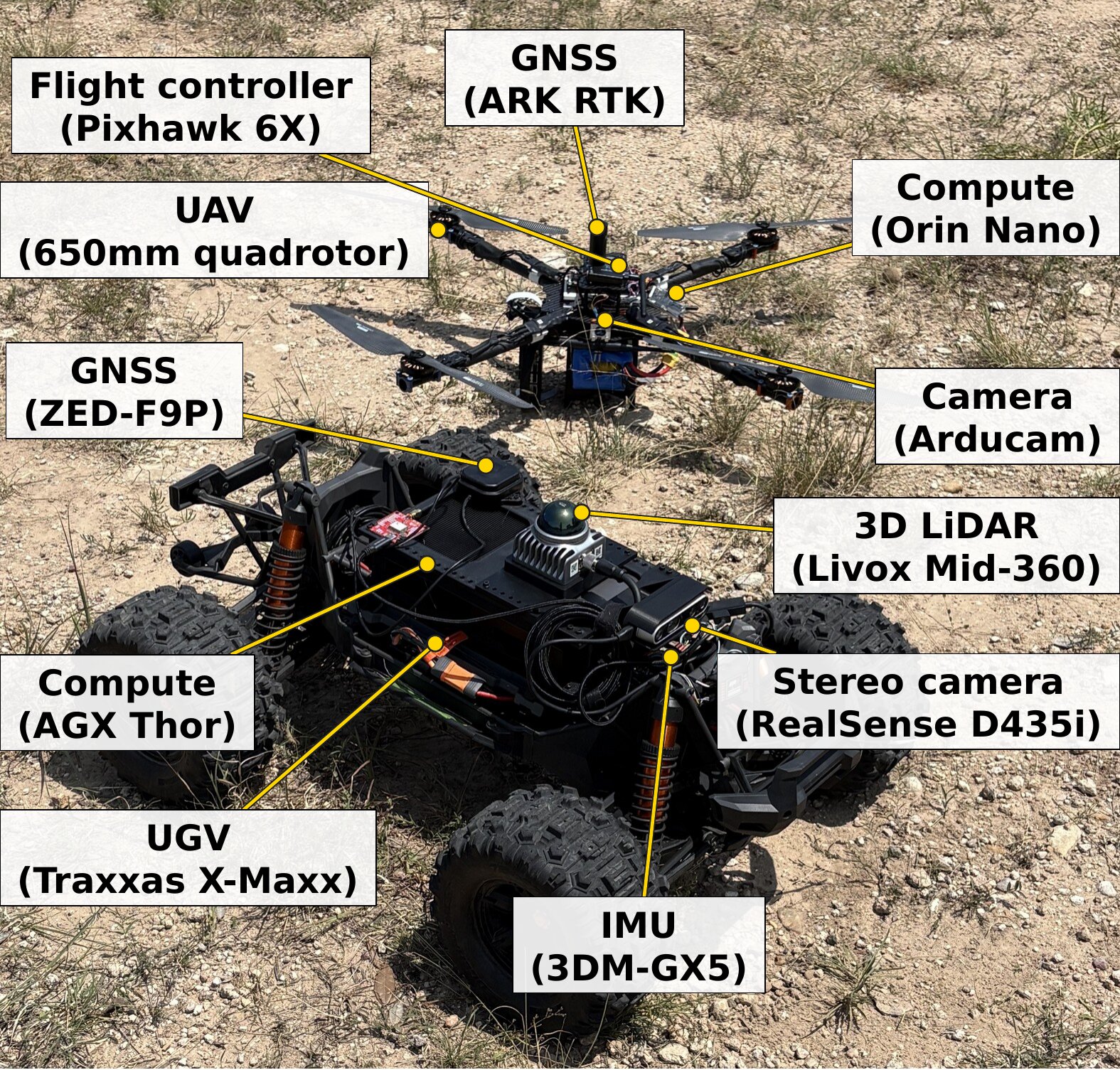}
    \caption{The Traxxas X-Maxx UGV and 650mm quadrotor UAV used in the field trials, with onboard sensing and compute labeled.
    \label{fig:field_platforms}}
\end{figure}

\begin{figure}[!b]
    \centering
    \includegraphics[width=0.85\linewidth]{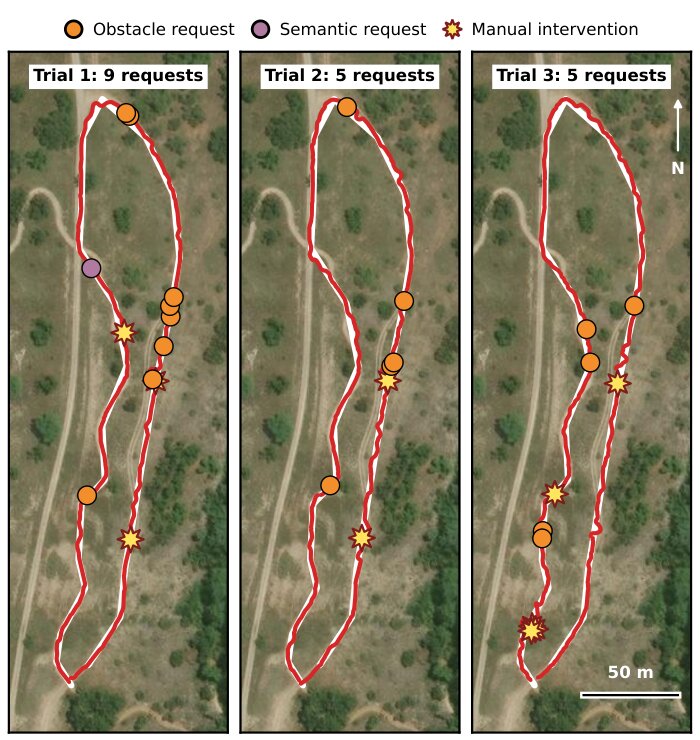}
    \caption{Ground Only baseline of the 563 m route by the autonomous UGV. The ground track is shown in red with markers indicating where the current uncertainty policy requested aerial refinement for obstacle uncertainty (orange) or semantic uncertainty (purple), and where manual intervention was required. The three trials produced 9, 5, and 5 candidate requests, respectively.
    \label{fig:field-routes}}
    \vspace{-.75em}
\end{figure}

\begin{figure*}[!t]
    \centering
    \includegraphics[width=0.9\linewidth]{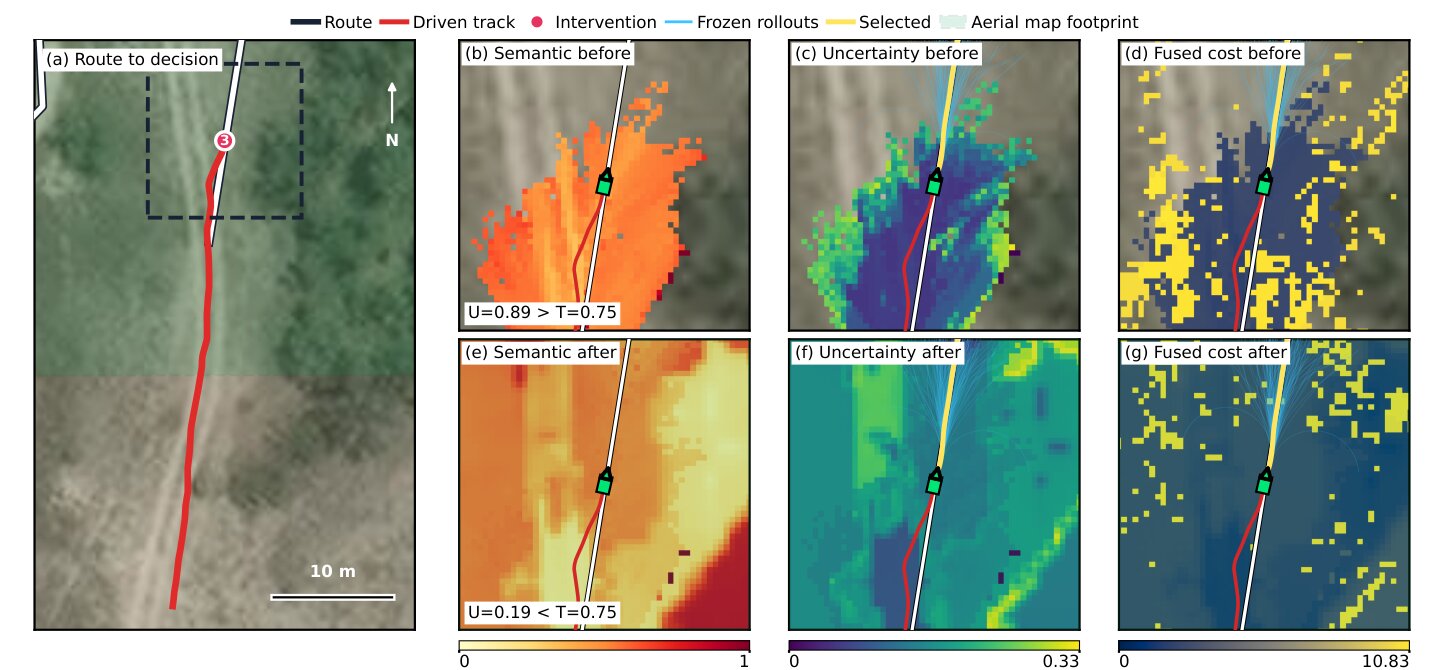}
    \caption{
Semantic map refinement following counterfactual assistance selection. The leftmost column shows the route segment where the counterfactual evaluation identified a need for aerial semantic mapping. The remaining three columns show the semantic map, its uncertainty, and the fused navigation cost map before and after incorporating the UAV-derived semantic map.}
    \label{fig:field-assisted-maps}
\end{figure*}

\section{Field Evaluation and Results}
To evaluate mission outcomes, aerial resource use, and navigation recovery under real-world uncertainty, we deployed the proposed counterfactual-assisted framework at an unstructured rural field site with hilly terrain, scattered shrubs and trees, tall grass, and pedestrian and vehicle trails.
\label{sec:field_results}

\textit{Field Autonomy.}
The UGV, labeled in Figure~\ref{fig:field_platforms}, is an Ackermann-steered off-road research platform built on a Traxxas X-Maxx chassis.
A motor equipped with a Hall-effect sensor and electronic speed controller provide propulsion and telemetry.
The sensing suite includes a MicroStrain 3DM-GX5-25 IMU, an Intel RealSense D435i RGB-D camera, and a Livox Mid-360 LiDAR with an integrated IMU.
We run ROS~2 on an NVIDIA Jetson Thor to process sensor data and control the UGV using the methods described in Section~\ref{sec:system_implementation}. We retain a radio-control interface 
for manual operation and safety intervention.
The UAV, labeled in Figure~\ref{fig:field_platforms}, is a quadcopter equipped with a Pixhawk 6X flight controller running PX4, a downward-facing Arducam HQ camera, and an ARK L1/L5 RTK GNSS module. An NVIDIA Jetson Orin Nano Super handles communication with the UGV, commands the flight controller during autonomous navigation, and generates the requested map layers.

The robots, along with an additional ground station, carry a Morse Micro MM8108 radio that supports communication in the 900~MHz band using IEEE 802.11ah Wi-Fi HaLoW \cite{adame2014ieee80211ah}.
The three devices connect over a mobile ad hoc network using the B.A.T.M.A.N.\ mesh-routing protocol \cite{neumann2008batman}, enabling long-range, multi-hop communication between network nodes.

For semantic mapping, the UGV uses CLIPSeg \cite{lüddecke2022image} on images from its downward-facing camera. Text embedding is grounded on terrain classes such as vegetation, trails, rocks, and water to provide traversability estimation.
When the UGV requests obstacle information, the UAV uses the DINOv3-based CHMv2 model to estimate the height above ground at each aerial image pixel \cite{brandt2026chmv2,simeoni2026dinov3}. Pixels with estimated height above ground exceeding 0.5 m are registered on the obstacle maps.
For both maps, the UAV geo-registers RGB input against an NAIP orthophoto using SuperPoint and SuperGlue correspondences followed by a RANSAC homography \cite{detone2018superpoint,sarlin2020superglue,fischler1981ransac} before sending them to the UGV.



\textit{Field Site and Evaluation Protocol.}
The UGV navigates the same predefined routes from Section~\ref{sec:simulation_results}. These routes cover changes in terrain, elevations, and physical obstacles. Similar to the simulation, we evaluate two navigation policies: (i) \textbf{Ground Only}, which uses onboard sensing without UAV assistance; and (ii) \textbf{UAV Counterfactual}, our proposed method.

We record the number of times the UGV engages in potentially hazardous behavior (attempting dangerous maneuvers, for example), the number of manual interventions (either radio control takeover or relocating the UGV), and the number of times it requests aerial assistance.
Figure~\ref{fig:field-routes} shows three trials of the Ground Only policy on the $\approx 563\textrm{ m}$ long route, and marks the locations where counterfactual assistance \emph{would} have been requested.
In the UAV Counterfactual runs, we record the number of times the UAV successfully responds with a map, and the number of times the aerial map resolves the uncertainty and UGV resumes navigation.

\textit{Unassisted and Assisted Performance.}
Across all field trials, the UGV autonomously covered approximately 2~km through challenging, unstructured terrain containing large bushes and dense grass.
These conditions limited ground-level perception and produced uncertain map cells along candidate UGV motions.
Ground only runs recorded three instances of risky behavior and required ten manual interventions.
Offline evaluation identified 19 locations where the assistance policy would have requested aerial evidence, comprising 18 obstacle-map requests and one semantic-map request.
During UAV Counterfactual runs, the system issued eight requests and successfully received six aerial maps; all six updates enabled the UGV to resume navigation. 
The two dropped requests were caused by network connectivity issues.
Figure~\ref{fig:field-assisted-maps} illustrates how aerial assistance resolved one such information deficit.
Vegetation along the route produced high semantic uncertainty, exceeding the 0.75 request threshold.
The returned aerial semantic map provided additional information about the surrounding terrain, updated the fused navigation costs, and reduced the uncertainty diagnostic to 0.19. With this refined representation, the UGV could continue navigation safely. These events demonstrate repeated use of aerial information to resolve navigation uncertainty in challenging field conditions.

\textit{Field Challenges.} Hardware reliability was the primary challenge in conducting consistent long-range navigation trials.
Communication was affected by interference among radios operating in the 900\,MHz band and by antenna placement and orientation.
We mitigated these issues by integrating a higher-gain omnidirectional antenna and adjusting its placement and alignment.
Communications equipment also overheated in direct sunlight, requiring additional cooling and improved airflow.
Although the UGV could traverse the rough terrain, grass fragments and dust accumulated in the onboard computer's cooling fans during repeated operation, restricting airflow and causing overheating; mesh filters were installed to reduce debris ingress. 
Maintaining UAV connectivity further highlighted the importance of antenna coverage in both azimuth and elevation.
Future experiments would benefit from a fixed, elevated antenna installation with a radiation pattern suited to the UAV's operating altitude.
Note that these failures occur on the hardware level rather than in counterfactual evaluations.

\section{Conclusion}
This paper presented a planner-conditioned air--ground assistance method that uses counterfactual map evaluations to decide when a UGV should request aerial sensing and what information the UAV should collect.
Across 180 simulation trials of the proposed method, it achieved 98.9\% goal attainment while reducing mean UAV flight time by 87.7\% relative to continuous greedy aerial mapping.
Aggregated across the three routes, each second of estimated UAV flight time saved $2.94\,\mathrm{s}$ of UGV navigation time relative to the ground only baseline, compared with $0.47\,\mathrm{s}$ for the greedy baseline.
Across approximately 2~km of unstructured autonomous UGV field operation, subsequent deployments demonstrated resolution of map uncertainty and navigation error.
Together, these findings show that targeted aerial sensing can preserve high goal attainment in simulation with substantially reduced aerial resource demand, while resolving information deficits encountered during navigation.
Future work will pursue long-range field missions through improved communication, recovery, and coordination mechanisms that increase reliability and reduce human intervention.

\bibliographystyle{IEEEtran}
\bibliography{refs}

\end{document}